\documentclass[letterpaper, 10 pt, conference]{ieeeconf}  

\IEEEoverridecommandlockouts                              

\let\labelindent\relax 
 
\usepackage{float}
\usepackage[latin]{babel}
\usepackage{blindtext}
\usepackage[usenames,dvipsnames]{xcolor}
\usepackage{graphicx}
\usepackage{stfloats}
\usepackage[inline]{enumitem}
\usepackage{wrapfig}
\usepackage[ruled]{algorithm2e}
\usepackage{tikz} 
\usepackage{bm}
\usepackage{siunitx}
\usepackage{amsmath}

\usepackage[para]{footmisc}

\makeatletter
\long\def\@makefntext#1{\leavevmode
    \@makefnmark\nobreak
    \hskip.1em\relax#1
}
\makeatother

\def\BibTeX{{\rm B\kern-.05em{\sc i\kern-.025em b}\kern-.08em
    T\kern-.1667em\lower.7ex\hbox{E}\kern-.125emX}}

\title{\LARGE \bf
Generating Future Observations to Estimate Grasp Success in Cluttered Environments
}

\author{Daniel Fernandes Gomes$^{1, 2}$, Wenxuan Mou$^{3}$, Paolo Paoletti$^{4}$ and Shan Luo$^{1}$
\thanks{$^{1}$Department of Engineering, King's College London, London WC2R 2LS, United Kingdom. Emails: \tt{\{danfergo, shan.luo\}@kcl.ac.uk}}
\thanks{$^{2}$smARTLab, Department of Computer Science, University of Liverpool, Liverpool L69 3BX, United Kingdom.}
\thanks{$^{3}$Department of Computer Science, University of Manchester, United Kingdom.
Email: \tt{wenxuan.mou@manchester.ac.uk}
}
\thanks{$^{4}$School of Engineering, University of Liverpool, Liverpool L69 3GH, United Kingdom. Email: \tt{paoletti@liverpool.ac.uk}}%
}

\begin{document}

\maketitle
\thispagestyle{empty}
\pagestyle{empty}

\begin{abstract}

End-to-end self-supervised models have been proposed for estimating the success of future candidate grasps and video predictive models for generating future observations. 
However, none have yet studied these two strategies side-by-side for addressing the aforementioned grasping problem.
We investigate and compare a model-free approach, to estimate the success of a candidate grasp, against a model-based alternative that exploits a self-supervised learnt predictive model that generates a future observation of the gripper about to grasp an object.
Our experiments demonstrate that despite the end-to-end model-free model obtaining a best accuracy of 72\%, the proposed model-based pipeline yields a significantly higher accuracy of 82\%. 
\textbf{https://danfergo.github.io/future-grasp}
\end{abstract}

\section{INTRODUCTION}


Several works have shown that Convolutional Neural networks can be trained in self-supervised manner to estimate the success of candidate gasps using setups that automatically label triplets of uncalibrated RGB scene observation,  robot command and grasp success \cite{HandEyeCoordination}. However, such end-to-end approaches require large datasets, that take days to collect and require costly hardware setups, while only being used for a single task. While these models can be trained for addressing multiple tasks \cite{garmnet}, exploring task-agnostic action-conditioned predictive models \cite{VisualForesight, maskvit} is a promising direction to improve data efficiency. Therefore, we investigate and compare these two scenarios wherein: 1) a single network is trained to predict the outcome of a candidate grasp, given the grasp configuration and the observation of the scene before the grasp is carried out; and 2) this pipeline is split into a predictive model, that predicts an observation taken right before the grasp occurs, and a grasp success estimator that outputs the grasp success, from such hallucinated observations. 

\begin{figure}
  \includegraphics[width=\linewidth]{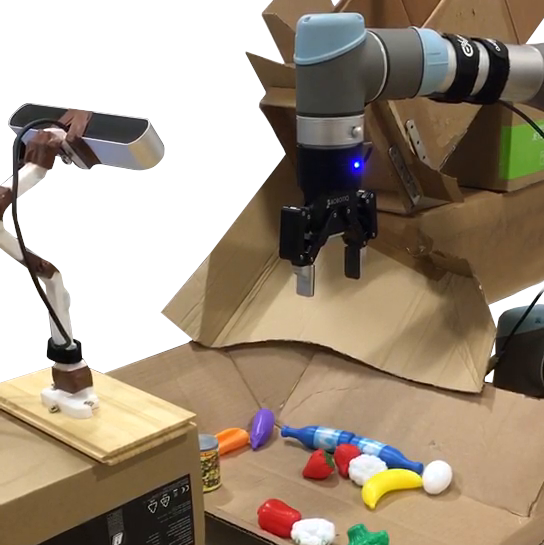}
  \caption{A robot arm randomly grasping various plastic objects from a bin, autonomously annotated using the gripper feedback.}
  \label{fig:cover}
\end{figure}

To optimize and evaluate our networks, a robot arm is setup to randomly grasp various plastic objects from a bin, with the collected data being autonomously annotated using the gripper feedback. A dispenser, near the bin, allows collecting grasps with a varied number of objects in the bin, as shown in Figure~\ref{fig:cover}. For each grasp, the setup saves 3 main moments: \textit{before} the robot and gripper are moved, \textit{during} the grasp (i.g., right before closing the gripper) and \textit{after} lifting the object. In total, we collect a dataset of 24,364 grasps over a period of 96h. We proceed with the training and analysis of the different networks, as presented in Section~\ref{sec:experiments}. 

\begin{figure*}[t]
  \includegraphics[width=\linewidth]{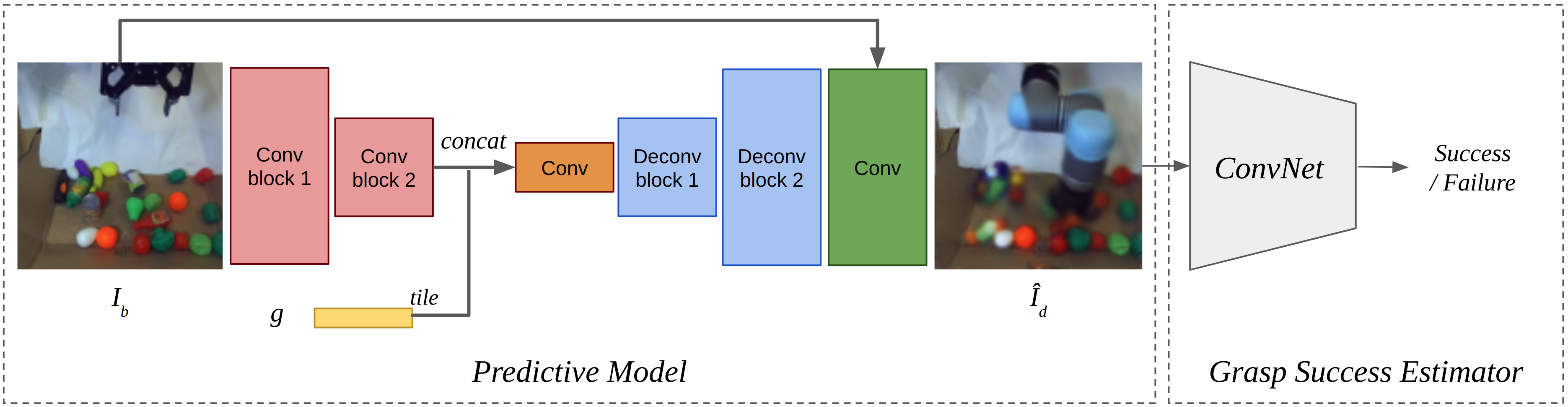}
  \vspace{-20pt}
  \caption{
  The predictive model generates the \textit{during} observation, $\hat{I}_{d}$, given the grasp command (gripper position, orientation and aperture) and \textit{before} observation, $I_{b}$. Then, the grasp success estimator, classifies this candidate grasp into successful or failure. 
  }
  \label{fig:method}
\end{figure*}

\section{EXPERIMENTS}
\label{sec:experiments}

To evaluate our hypothesis experiments are carried out in three stages. Firstly, a model-free estimator of the grasp success, given  the \textit{before} observation and grasp command, is trained. Then, the same estimator is obtained given the observation \textit{during} the grasp. Finally, a predictive model is trained to generate the future \textit{during}, from the \text{before} observation and grasp command, followed by a discriminator that estimates the success of such grasp.


\subsection{Challenges with model-free grasp success prediction}
\label{sec:exp_a}
We start with training a standard ConvNet to directly predict the grasp success from the \textit{before} observation and grasp command. We carry out extensive experiments while varying training parameters and approaches for combining the grasp command with the remaining image features.
However, in all scenarios either the network either overfits the training data, and fails to generalize to the validation split, or plateaus from the beginning of the training (this occurs particularly when heavy data augmentation is used). In the best case, an accuracy of 72\% is achieved.



\subsection{Predicting success when the grasp occurs}
\label{sec:exp_b}

Predicting the outcome of a candidate's grasp comes down to reasoning over the relationship between the gripper and the object to be grasped right before the grasp takes place. 
Therefore, we hypothesise that estimating grasp success from the observation of the gripper about to grasp the object, should result in a higher accuracy. After filtering cases where the gripper highly occludes the in-grasp object, i.e. with the arm wrist having an angle between \SI{-0.75}{\radian} and \SI{0.75}{\radian}, the network obtains a mean accuracy of $89.7\%$. While this model can not be used in the real world, it serves as an upper bound for the results of the following experiment.

\begin{table}[]
\centering
\caption{Grasp success probability estimator accuracy}
\def\arraystretch{1.3}
\begin{tabular}{l|l|l}
\hline
& $\uparrow$ \textbf{Train} 
& $\uparrow$ \textbf{Validation} 
\\
\hline
model-free, $p(s | I_b, c)$ & $72\%$ & $72\%$  \\ \hline
surrogate, $p(s | I_d)$ & \bm{$85.4\%$} & \bm{$89.7\%$} \\ \hline
model-based, $f(I_b, c) \rightarrow \hat{I}_d$,  $p(s | \hat{I}_d)$ &  $82.3\%$ & $82.0\%$ \\ \hline


\end{tabular}
\label{table:data_collection}
\end{table}

\begin{figure}
\centering
\includegraphics[width=\linewidth]{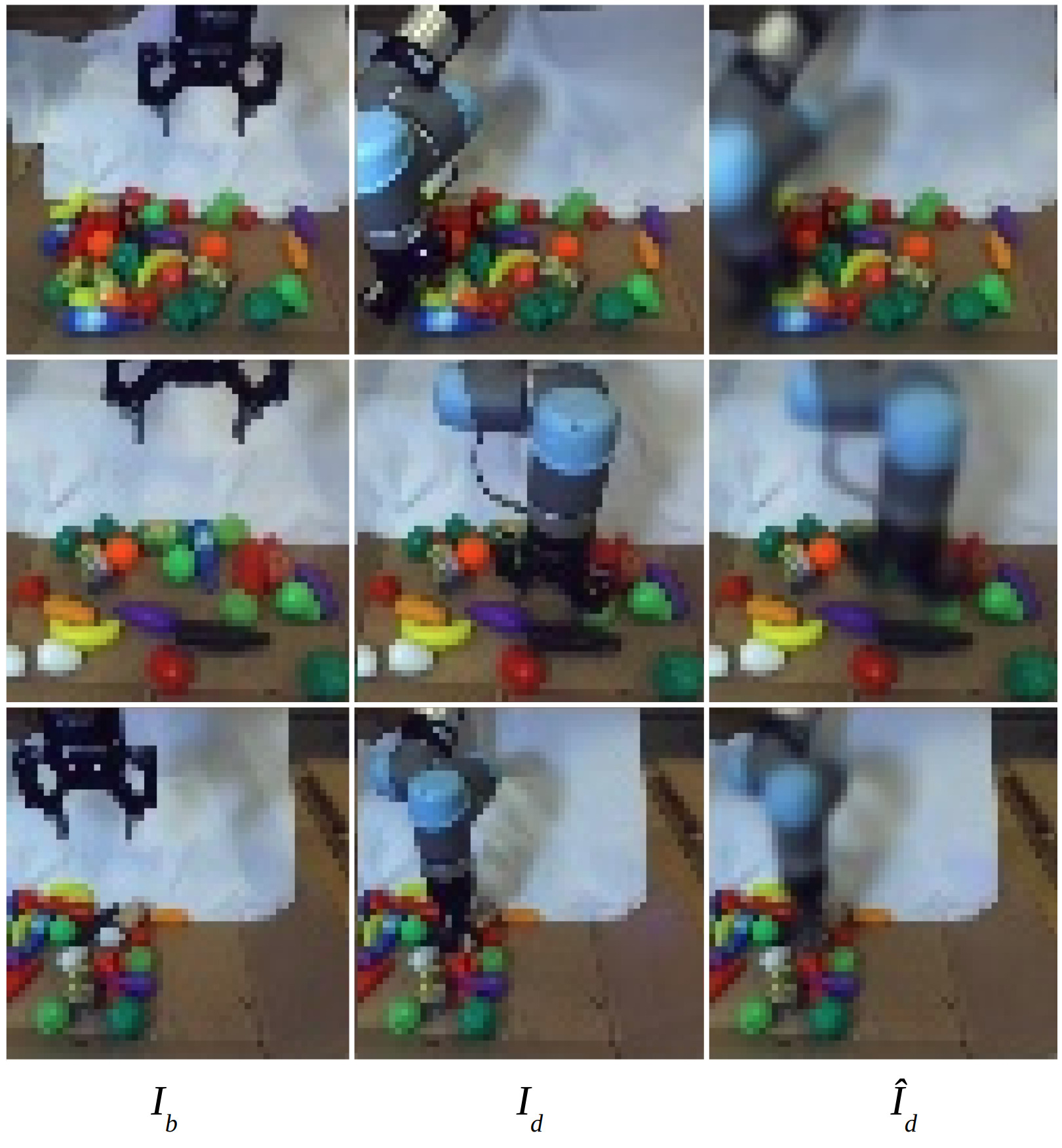}
\caption{\textit{During} observations generated by the predictive model, given the \textit{before} observation and arm configuration. While the background sharpness is highly due to the skip connection, as proposed in \cite{VisualForesight}, we note the model capacity of  omiting the gripper in \textit{before}.}
\label{fig:predictive_model_samples}
\end{figure}

\subsection{Predicting grasp success from hallucinated observations}
\label{sec:exp_d}

Finally, we train a predictivel model to generate the \textit{during} observations, from the \textit{before} and action command. Then, these generated observations are used to train the grasp success estimator, as shown in Figure~\ref{fig:method}.  Whilst still yielding blurry hallucinations, the predictive model is able to render the arm and gripper in the correct pose, as shown in Figure~\ref{fig:predictive_model_samples}. With that, the combined pipeline achieves an accuracy of $82.0\%$, which constitutes a significant improvement (10\%) over the model-free baseline (\ref{sec:exp_a}).




\section{Conclusions and Future Work}

This work demonstrates the importance of considering future-prediction models, that can be trained end-to-end and without requiring human annotations. Leveraging the hallucinated views of the scene, the model-based pipeline achieves significantly better results than the model-free alternative. Future work should look into improving the generated future observations, using the more recent diffusion and transformer architectures and onto extending this approach to include optical tactile sensing in Sim2Real setting \cite{gomes2023flat, geltip, gelsightSim}.

\bibliographystyle{IEEEtran}
\bibliography{references.bib}

\end{document}